\documentclass[10pt,twocolumn,letterpaper]{article}

\usepackage[pagenumbers]{cvpr} 
\usepackage{multirow}
\usepackage{multicol}
\usepackage{pifont}
\usepackage{tabularx}
\usepackage[dvipsnames]{xcolor}

\definecolor{cvprblue}{rgb}{0.21,0.49,0.74}
\usepackage[pagebackref,breaklinks,colorlinks,citecolor=cvprblue]{hyperref}

\def\paperID{*****} 
\def\confName{CVPR}
\def\confYear{2024}

\title{Scaling Continuous Kernels with Sparse Fourier Domain Learning}

\author{
    Clayton Harper$^{1}$, Luke Wood$^{2}$, Peter Gerstoft$^{2}$, Eric C. Larson$^{1}$ \\
    $^1$Southern Methodist University \\
    $^2$University of California San Diego \\
    {\tt\small caharper@smu.edu, lukewoodcs@gmail.com, pgerstoft@ucsd.edu, eclarson@smu.edu}
}

\begin{document}
\maketitle
\begin{abstract}

We address three key challenges in learning continuous kernel representations: \textit{computational efficiency}, \textit{parameter efficiency}, and \textit{spectral bias}. Continuous kernels have shown significant potential, but their practical adoption is often limited by high computational and memory demands. Additionally, these methods are prone to spectral bias, which impedes their ability to capture high-frequency details.  To overcome these limitations, we propose a novel approach that leverages sparse learning in the Fourier domain. Our method enables the efficient scaling of continuous kernels, drastically reduces computational and memory requirements, and mitigates spectral bias by exploiting the Gibbs phenomenon.
\end{abstract}    
\section{Introduction}
\label{sec:intro}

Continuous kernel representations have emerged as a powerful method to learn large convolutional kernels with a fixed-parameter budget \cite{romero2022ckconv,romero2022flexconv,knigge2023modelling,romero2022learning,romero2023dnarch}. Instead of utilizing a discretely parameterized CNN kernel, a small neural network, typically a multi-layer perceptron (MLP), generates the convolutional kernel by sampling the MLP at various spatial positions. In this formulation, convolutional kernels are modeled as continuous, vector-valued functions, allowing them to be sampled for arbitrary resolutions.  By querying the MLP at more densely spaced positions, the size of the convolutional kernel can be effectively increased without adding additional parameters. 

Continuous kernel formulations not only offer greater flexibility but also allows networks to dynamically learn the effective kernel sizes through gradient-based optimization. Other techniques have explored gradient-based methods for learning convolutional kernel sizes dynamically \cite{8237351, 7780655, 9552550}. However, many of these methods are constrained by their reliance on a single kernel size, applied uniformly across all channels and filters within a layer. This homogeneity can limit the network's capacity to capture diverse spatial dependencies.  Continuous kernel representations, in contrast, provide a more flexible approach by generating kernels across channels. Despite these advantages, the practical deployment of continuous kernels is hampered by three key challenges: (1) high parameter counts, (2) high computational and memory demands during training, and (3) the susceptibility to spectral bias, which limits their ability to capture high-frequency components in the data. 

First, the computational overhead of training continuous kernels is substantial. Since the entire convolutional kernel must be generated on-the-fly during each forward pass, this process requires significant resources. In contrast, traditional CNNs directly learn discrete weights to be applied. Furthermore, due to the use of automatic differentiation in modern frameworks like PyTorch and TensorFlow, continuous kernel methods necessitate large amounts of memory, as intermediate activations must be stored for gradient computation. This makes scaling to large-scale applications prohibitively expensive. 

Second, spectral bias \cite{basri2020frequency,rahaman2019spectral} presents a fundamental limitation. Neural networks tend to favor low-frequency components, resulting in poor generalization when fine-grained details or high-frequency information is required. This bias is particularly problematic in tasks where the ability to model sharp transitions or high-frequency variations is critical.

In this work, we propose a novel approach, \textit{Continuous Fourier Convolutions} (CF-Convs), which addresses these challenges by learning continuous kernel representations in the Fourier domain. By sparsely updating the generated kernels, we significantly reduce computational costs and memory usage, leading to faster and more efficient training. Furthermore, by learning in the Fourier domain, we exploit the Gibbs phenomenon to mitigate spectral bias, ensuring that our learned kernels capture a broader and more balanced frequency spectrum.

\begin{figure*}[t!]
    \centering
    \includegraphics[width=\linewidth]{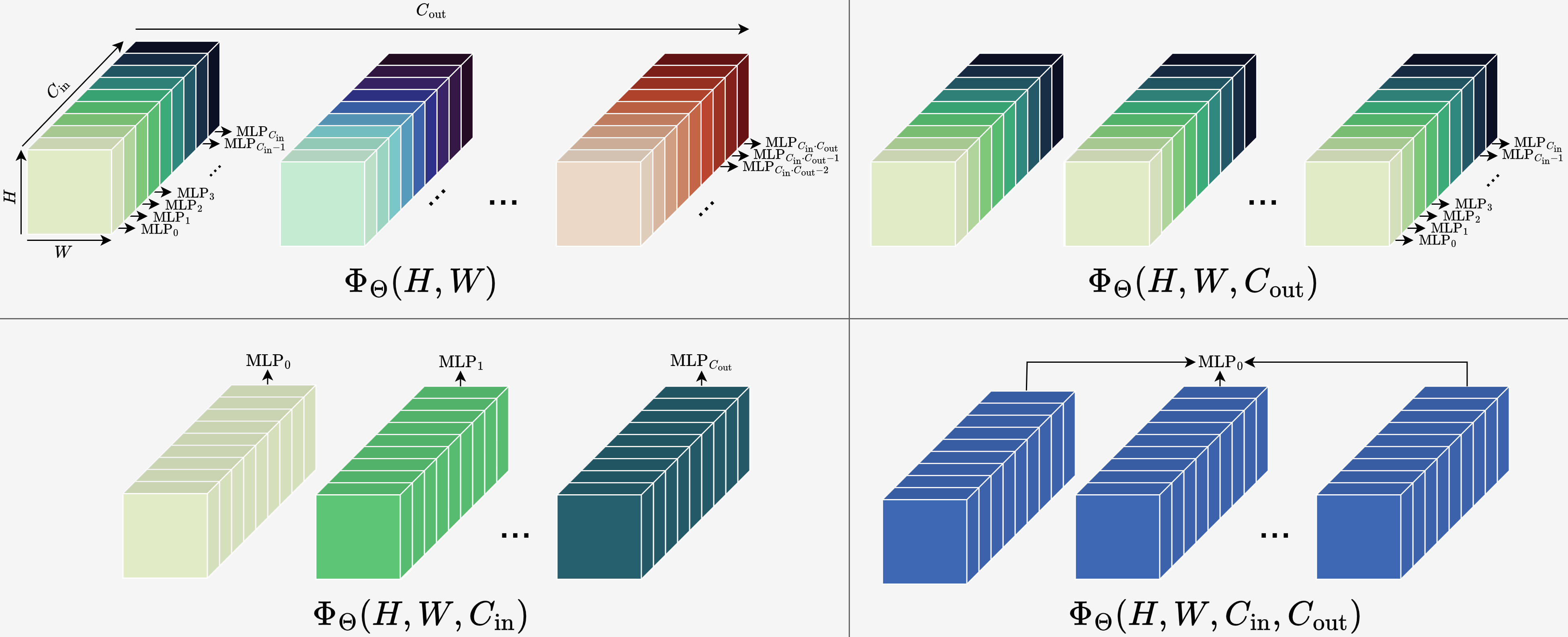}
    \caption{Illustration of different potential parameterizations for CF-Conv layers, where MLPs are conditioned on various axes. The number of MLPs (represented by different colors) and the corresponding parameter counts vary based on the chosen parameterization.}
    \label{fig:if_conv_parameterizations}
\end{figure*}

Our main contributions are as follows:

\begin{enumerate} 
    \item Our method, CF-Convs learn directly in the Fourier domain, mitigating the risks of spectral bias through leveraging the Gibbs phenomenon.
    \item Using a continuous kernel representation, CF-Convs avoid parameter explosion associated with Fourier domain learning.
    \item To tackle the computational and memory inefficiencies of continuous kernel learning, we introduce a sparse update mechanism that accelerates training and reduces memory consumption.
\end{enumerate}

\section{Related Work}
Recent architectures have leveraged continuous representations to parameterize convolutional kernels as vector-valued functions, enabling adaptive sampling and kernel size scaling without parameter explosion \cite{romero2022ckconv, romero2022flexconv, knigge2023modelling, romero2022learning, romero2023dnarch}. This paradigm draws on the broader concept of implicit neural representations, as seen in works such as \cite{mescheder2019occupancy, park2019deepsdf, sitzmann2020implicit}, where neural networks encode continuous signals. By allowing convolutional kernels to adapt dynamically to spatial patterns in the data, these methods enhance the network's expressiveness.

Most existing approaches, however, have focused on learning in the spatial domain. A key exception is the work by Wood and Larson, which introduced parameterized techniques in the Fourier domain. They proposed using 2D Gaussian parameterized kernels to reduce the parameter costs typically associated with Fourier domain learning \cite{wood2021parametric}. While this approach facilitated optimization within the Fourier domain, effectiveness was curtailed due to shared weights across all input channels. Consequently, the Fourier kernels exhibited uniform spatially-equivalent kernel sizes across channels, similarly to \cite{8237351,7780655,9552550}.

Despite its limitations, the Fourier domain offers key advantages over spatial domain approaches.  Our work builds upon these Fourier-based techniques by addressing the shortcomings in prior methods.
\section{Fourier Domain Motivation}

    \definecolor{darkyellow}{rgb}{0.74, 0.68, 0.11}
    \newcommand{\vlow}{\textcolor{green}{\(\bullet\)}}
    \newcommand{\low}{\textcolor{darkyellow}{\(\bullet\bullet\)}}
    \newcommand{\medium}{\textcolor{orange}{\(\bullet\bullet\bullet\)}}
    \newcommand{\high}{\textcolor{red}{\(\bullet\bullet\bullet\bullet\)}}

    \renewcommand{\arraystretch}{1.3}
    \begin{table*}[h!]
    \caption{Comparison of different parameterization methods. Arrows indicate direction of better values.  Memory usage refers to GPU memory consumption during training, not inference. Memory usage was evaluated using a 6-layer CNN with 32 filters per layer, applied to an input of size $150 \times 150 \times 3$. This network corresponds to the architecture in \Cref{fig::cf-conv-arch}. The number of MLPs is doubled to account for the real and imaginary components of the split kernel.}
        \centering
        \begin{tabular}{@{} lccc p{4cm} @{}}
        \toprule
        \textbf{Parameterization} & \textbf{\# of MLPs} & \textbf{\# Params $\downarrow$} & \textbf{Memory $\downarrow$} & \textbf{Notes*} \\
        \midrule
        Spatial $3 \times 3$ CNN & --- & \vlow & \vlow & Fits on 80GB GPU \\ \hline
        $\Phi_\Theta(H, W)$ & $C_{\text{in}} \cdot C_{\text{out}} \cdot 2$ & \high & \low & Fits on 80GB GPU \\ \hline
        $\Phi_\Theta(H, W, C_{\text{in}})$ & $C_{\text{out}} \cdot 2$ & \medium & \medium & Naive implementation exhausts 80GB of GPU memory \\ \hline
        $\Phi_\Theta(H, W, C_{\text{out}})$ & $C_{\text{in}} \cdot 2$ & \medium & \medium & Naive implementation exhausts 80GB of GPU memory \\ \hline
        $\Phi_\Theta(H, W, C_{\text{in}}, C_{\text{out}})$ & $1 \cdot 2$ & \vlow & \high & Naive implementation exhausts 80GB of GPU memory \\
        \bottomrule
        \end{tabular}
        \label{table:parameterization}
    \end{table*}

    \label{sec::fourier_motivation_1}

    \textbf{Terminology}: Let $H$ and $W$ represent the image height and width, respectively. $H_k$ and $W_k$ are the height and width of the kernel, while $C_{\text{in}}$ and $C_{\text{out}}$ denote the number of input channels and output channels (filters).

    MLPs are known to struggle with generating high-frequency functions, an issue known as ``spectral bias'' \cite{basri2020frequency,rahaman2019spectral}.  To remedy this issue, previous works have proposed the use of random Fourier features \cite{rahimi2007random,sutherland2015error,tancik2020fourier} and sinusoidal representation networks (SIRENs) \cite{sitzmann2020implicit}.  In the context of continuous kernel representations, both Fourier features and SIRENs have demonstrated substantial promise \cite{romero2022flexconv,romero2022ckconv,van2022relaxing}.  Building on these insights, we propose that learning directly in the Fourier domain can further enhance performance.

    Spectral bias in the spatial domain hinders the ability to learn high-frequency components, which are critical for capturing fine-grained details. By learning in the Fourier domain, we can alleviate this issue using the Gabor limit. Energy concentrated in one domain results in a wide distribution in its reciprocal domain, meaning that low-frequency functions in the Fourier domain can correspond to high-pass filters in the spatial domain. For instance, a smooth kernel in the Fourier domain (i.e., generated from a low-frequency function) can concentrate energy in high-frequency regions.  This duality allows for Fourier representations to learn high-pass filters, mitigating the effects of spectral bias.

    However, learning discretely in the Fourier domain introduces a substantial increase in parameter count, which scales with the input size. Specifically, the number of parameters in a convolutional layer is $H_k \cdot W_k \cdot C_{\text{in}} \cdot C_{\text{out}}$. In the Fourier domain, $H_k$ and $W_k$ match the image dimensions ($H$ and $W$), leading to a significant increase in parameter cost. For example, consider a CNN with 10 input channels and 32 filters, applied to $150 \times 150$ images. A spatial-domain kernel with size $3 \times 3$ would require $3 \cdot 3 \cdot 10 \cdot 32 = 2,880$ parameters. In the Fourier domain, an \textit{equivalent} kernel would require $150 \cdot 150 \cdot 10 \cdot 32 \cdot 2 \approx 14.4$M parameters (the factor of 2 accounts for the complex-valued representation). This dramatic increase is a key limitation in the discrete Fourier domain.

    To address both spectral bias and the issue of parameter explosion, we propose learning continuous kernels in the Fourier domain. Our approach, Continuous Fourier Convolutions (CF-Convs), maintains the advantages of frequency-based learning while avoiding the excessive parameter overhead associated with discrete Fourier representations.

\section{CF-Conv Parameterizations}

A convolutional layer in the Fourier domain for a given layer $L$ is defined as:
\begin{align}
    G^{(L)} = F \star K = \sum_{i=1}^{C_{\text{in}}} F_{i} \odot K_{i, o}, \quad \forall o \in C_{\text{out}}
\end{align}
where $F \in \mathbb{C}^{(H \times W \times C_{\text{in}})}$ is the input feature map, $K \in \mathbb{C}^{(H \times W \times C_{\text{in}} \times C_{\text{out}})}$ is the convolutional kernel, $G \in \mathbb{C}^{(H \times W \times C_{\text{out}})}$ is the output, and $\odot$ denotes element-wise multiplication of complex values. Due to the the Fourier transform producing a complex-valued output, the convolutional layer must separately learn real, $K_{i,o}^R$, and imaginary, $K_{i,o}^I$, kernel components.  This configuration, known as a split kernel configuration \cite{8495012}, is represented as $K_{i, o} = K_{i,o}^R + j\,K_{i,o}^I$, where $K_{i,o} \in \mathbb{C}^{(H \times W)}$.

Continuous kernels are generated by sampling parameterized functions at specified positions. Various forms of sampling are possible, as illustrated in \Cref{fig:if_conv_parameterizations}. MLPs can be conditioned on different axes ($H$, $W$, $C_{\text{in}}$, or $C_{\text{out}}$), leading to different parameterizations that affect both the number of MLPs required and the overall parameter count. To generate the complete convolutional kernel, $H \cdot W \cdot C_{\text{in}} \cdot C_{\text{out}}$ positions must be evaluated in each forward pass. Each parameterization offers a different balance of flexibility, memory usage, and computational complexity, as outlined in \Cref{table:parameterization}.

    Memory usage in all CF-Conv configurations is substantial due to the auto-differentiation process, where intermediate activations must be stored to compute gradients. Since the entire convolutional kernel is generated at each forward pass, the gradient must be computed for each $H \times W \times C_{\text{in}} \times C_{\text{out}}$ position. 

    In some parameterizations, memory can be conserved by freeing intermediate activations that are not required for later computations. For instance, in the $\Phi_\Theta(H, W, C_{\text{in}})$ configuration, each filter is independent, allowing memory to be released after summing over $C_{\text{in}}$. However, in the $\Phi_\Theta(H, W, C_{\text{in}}, C_{\text{out}})$ approach, a single MLP generates the entire kernel, requiring gradients to be computed over all positions. Therefore, memory can not be as readily freed, resulting in larger memory utilization. Essentially, this approach is trained on a `batch size' of $H \cdot W \cdot C_{\text{in}} \cdot C_{\text{out}}$.  

    \begin{figure*}[t]
    \centering
    \includegraphics[width=\linewidth]{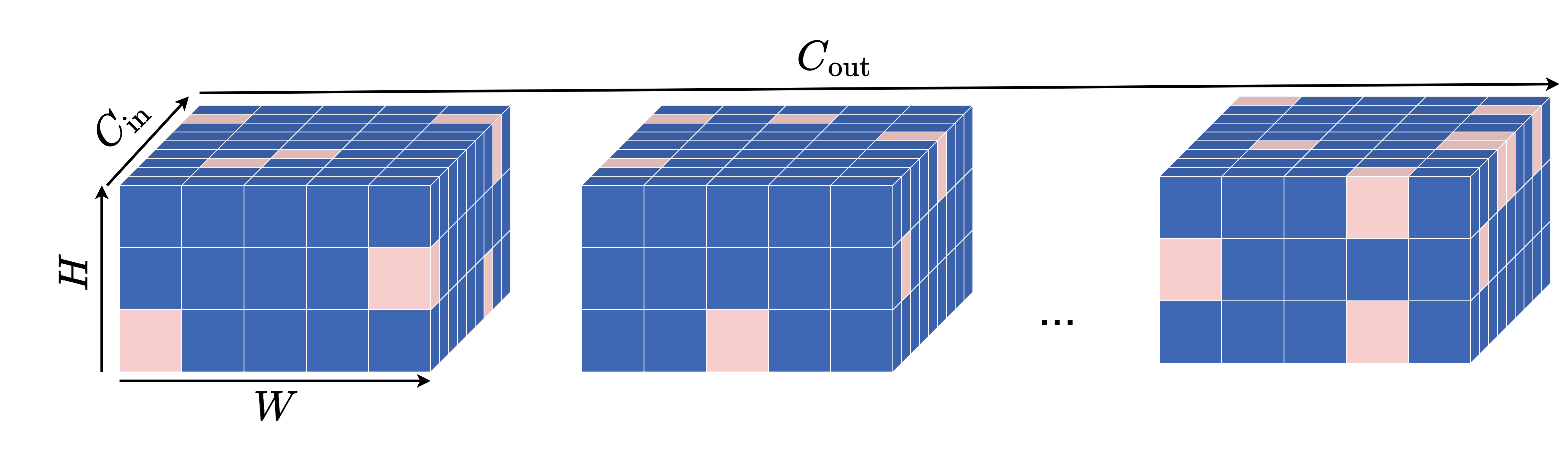}
    \caption{Sparse sampling visualization for more efficient training of CF-Conv layers.  Uniformly random sampled positions for a single training step are shown in red.}
    \label{fig::mlp_sampling}
\end{figure*}

    The ideal parameterization should offer sufficient expressivity with the fewest possible parameters. For these reasons, we advocate for the adoption of the $\Phi_\Theta(H, W, C_{\text{in}}, C_{\text{out}})$ configuration, which is similar to the approach in \cite{romero2022ckconv}. This setup offers an attractive trade-off, requiring only one MLP, which can be easily scaled by increasing its depth or width to handle more complex tasks without a significant increase in the number of trainable parameters. Although alternative parameterizations may offer reduced memory usage, they result in a substantial increase in parameters due to the need for multiple MLPs. To fully harness the potential of the $\Phi_\Theta(H, W, C_{\text{in}}, C_{\text{out}})$ approach, addressing memory and computational inefficiencies is critical to its practical implementation.
\section{Scaling CF-Convs}

\begin{figure*}[t]
    \centering
    \includegraphics[width=.75\linewidth]{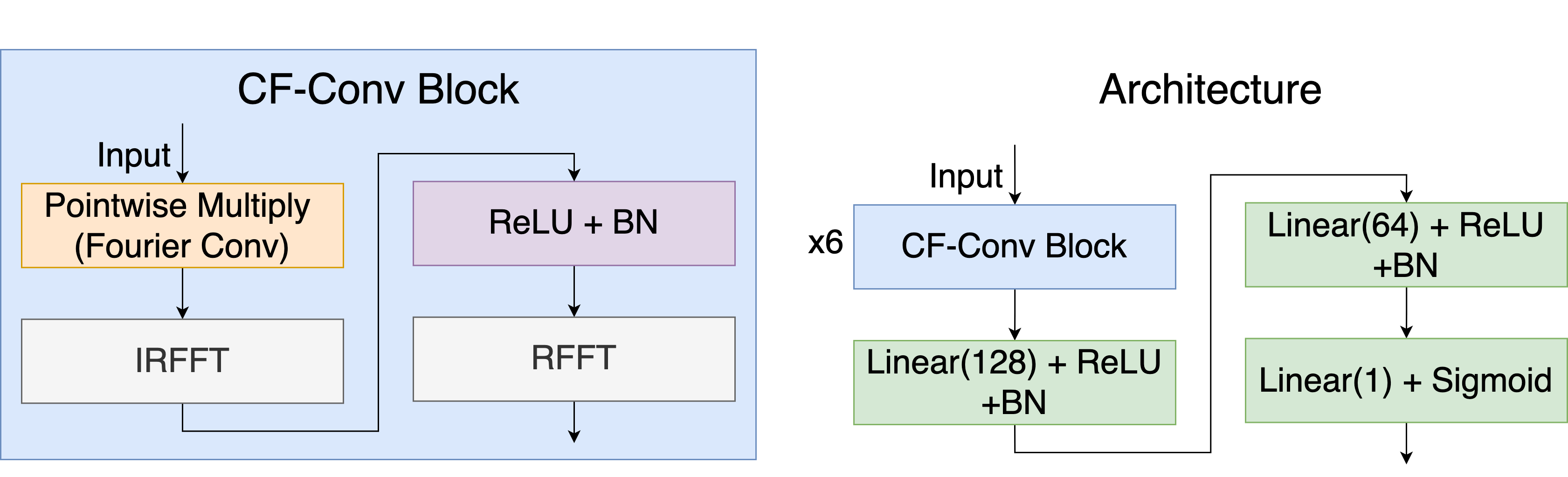}
    \caption{Cats vs. Dogs architectural overview using CF-Convs.}
    \label{fig::cf-conv-arch}
\end{figure*}

Several strategies can be employed to increase efficiency of CF-Convs, including gradient checkpointing, scan operations, and exploring more memory-efficient kernel representations. 
The core challenge in scaling the $\Phi_\Theta(H, W, C_{\text{in}}, C_{\text{out}})$ approach stems from the effective `batch size' of $H \cdot W \cdot C_{\text{in}} \cdot C_{\text{out}}$. This results in a significant number of intermediate activations that need to be stored during backpropagation.

One approach to reduce memory consumption during backpropogation is the use of gradient checkpointing, or rematerialization \cite{chen2016training,kumar2019efficient}. Instead of storing all intermediate activations required for gradient computations, only a subset is stored, and the remaining activations are recomputed during the backward pass as needed. This reduces memory usage at the expense of additional computation during backpropagation, as the activations must be recomputed.

Another method of reducing memory complexity is the use of a \textit{scan} operation.  This approach applies a given function sequentially over a collection of elements, accumulating intermediate results rather than storing all values simultaneously.  This effectively creates mini-batches and processes the $H \times W \times C_{\text{in}} \times C_{\text{out}}$ positions in chunks rather than all at once like the naive implementation.  Additionally, scan operations can be unrolled to control the size of the mini-batch.

We leverage both gradient checkpointing and reducing scan operations to allow the $\Phi_\Theta(H, W, C_{\text{in}}, C_{\text{out}})$ approach to fit on an 80GB GPU using a 6-layer, 32 filter CNN on image sizes of $150 \times 150 \times 3$. However, these methods act more as analgesics, addressing the symptoms rather than solving the core problem of high memory consumption. In the context of CF-Convs, these approaches are particularly impractical due to the significant increases in training times as shown in \Cref{tab::epoch_training_time}. 
    
\begin{table}[h]
    \caption{Epoch training time comparison for various methods on the Cats vs. Dogs dataset, utilizing a 6-layer CNN with 32 filters per layer.  All training times are reported using A100 80GB GPUs.}
    \centering
    \begin{tabular}{>{\raggedright}m{4cm} c}
        \toprule
        \textbf{Method} & \textbf{Epoch Training Time} \\
        \midrule
        Spatial $3 \times 3$ CNN & $\sim 30$s \\
        $\Phi_\Theta(H, W, C_{\text{in}}, C_{\text{out}})$ & \\
        \quad Naive & Exhausts 80GB GPU RAM \\
        \quad Rematerialization & $>2$ days \\
        \quad Scan & $>2$ days \\
    
    \end{tabular}
    \label{tab::epoch_training_time}
\end{table}

\subsection{Sparse Updates}

    Due to the impracticality of the aforementioned approach, we propose a training method that simultaneously decreases memory utilization while improving training speed. Our method consists of two principle components: 1) sparse evaluations to reduce memory consumption and training times 2) sparse updates to increase training stability.
    
    To reduce memory consumption, which arises from the need to evaluate the MLP at a large number of positions, we compute the gradient over a sampled subset of kernel positions. We denote these positions as \textit{selected positions}. As illustrated in \Cref{fig::mlp_sampling}, these positions (highlighted in red) are randomly sampled at each training step. We employ uniform sampling to ensure no position is given priority. This sparse evaluation approach dramatically reduces memory consumption, as fewer activations must be stored for the gradient update. Additionally, sparse evaluations reduce training time as fewer computations are required.

    To ensure stability during training, we employ stateful variables and sparse updates, similar to batch normalization's handling of batch statistics \cite{batch_norm}. Upon initialization, a kernel of shape $H \times W \times C_{\text{in}} \times C_{\text{out}}$ is stored as a state variable for both the real and imaginary components of the CF-Conv layer. The kernel is randomly initialized with a uniform distribution in the range of [-1,1], ensuring no bias towards any specific spatial position or frequency.
    
    During training, the kernel is updated at the selected positions using an exponential moving average (EMA) with a decay rate of $0.1$. This EMA approach smooths the updates and reduces the influence of outlier gradients, promoting stable learning dynamics. Additionally, the use of stateful variables allows the \textit{unselected positions} (highlighted in blue in \Cref{fig::mlp_sampling}) to evolve over time, even though they are not directly updated during each iteration.

         \begin{table}[h]
    \caption{Epoch training time comparison for various methods on the Cats vs. Dogs dataset, utilizing a 6-layer CNN with 32 filters per layer.  All training times are reported using A100 80GB GPUs.}
    \centering
    \begin{tabular}{>{\raggedright}m{4cm} c}
        \toprule
        \textbf{Method} & \textbf{Epoch Training Time} \\
        \midrule
        Spatial $3\times 3$ CNN & $\sim 30$s \\
        $\Phi_\Theta(H, W, C_{\text{in}}, C_{\text{out}})$ & \\
        \quad Naive & Exhausts 80GB GPU RAM \\
        \quad Rematerialization & $>2$ days \\
        \quad Scan & $>2$ days \\
        \quad Sparse updates &  \\
        \quad \quad +Scan & \\
        \quad \quad \quad \quad $2^{18}$ & $\sim 18$ min \\
        \quad \quad +Vmap & \\
        \quad \quad \quad \quad $2^{12}$ & $\sim 4.5$ min \\
        \quad \quad \quad \quad $2^{15}$ & $\sim 4.5$ min \\
        \quad \quad \quad \quad $2^{18}$ & $\sim 5$ min \\
        \quad \quad \quad \quad $2^{21}$ & $\sim 8$ min \\
        \bottomrule
    \end{tabular}
    \label{tab::epoch_training_time_updated}
\end{table}

    We experimented with different numbers of selected positions—$2^{12}$, $2^{15}$, $2^{18}$, and $2^{21}$—to evaluate the trade-off between memory consumption and training stability. As shown in \Cref{tab::epoch_training_time_updated}, sparse updates combined with scan or vmap operations dramatically reduce training time compared to the rematerialization and naive+scan approaches. Using vmap (vectorized map) further improves efficiency by running parallel computations, whereas scan performs these operations sequentially. With a small enough number of selected positions, vmap-based methods can fit into memory and offer significant speed advantages.

    \begin{table*}[t]
    \caption{Number of parameters and accuracy for different CF-Conv parameterization methods. Best results are shown in bold.}
    \centering
    \begin{tabular}{l c c c}
        \toprule
        \textbf{Method} & \textbf{\# Params} $\downarrow$ & \textbf{Accuracy (\%) $\uparrow$} & Notes \\
        \midrule
        Spatial $3 \times 3$ CNN & 60K & \textbf{86.64} & --- \\
        $\Phi_\Theta(H, W)$ & 107K & 83.53 & --- \\
        $\Phi_\Theta(H, W, C_{\text{in}}, C_{\text{out}})$ naive & 59K & --- & Exhausts 80GB GPU RAM \\
        $\Phi_\Theta(H, W, C_{\text{in}}, C_{\text{out}})$ w/ sparse updates & &  \\
        \quad +Vmap & & & --- \\
        \quad \quad \quad $2^{12}$ & 59K & 75.26 & --- \\
        \quad \quad \quad $2^{15}$ & 59K & 79.27 & --- \\
        \quad \quad \quad $2^{18}$ & 59K & 85.30 & --- \\
        \bottomrule
    \end{tabular}
    \label{tab::sparse_params_accuracy}
\end{table*}

    While there remains an approximately 10$\times$ difference in training speeds compared to spatial CNNs, several important considerations contextualize this disparity. Firstly, the comparison is against spatial CNNs utilizing small $3 \times 3$ filters, whereas our layers can learn kernels ranging from spatial equivalents of $1 \times 1$ to $H \times W$ in about 5 minutes. Training time for our layers is agnostic to the kernel size, taking the same 5 minutes whether the spatial equivalent is $1 \times 1$ or $H \times W$. In contrast, spatial CNNs experience increased training times with larger kernel sizes.  In the worst case scenario where the kernel is $H \times W$ the algorithmic complexity of performing the forward pass of a single kernel is ${H \times W}^2$ whereas with our approach the algorithmix complexity remains $H \times W$.  As such, while a performance gap continues to exists for small sized kernels our CNNs are able to efficiently scale to arbitrarily sized kernels.
\section{Experiments and Results}

To evaluate the performance of our CF-Convs, we employ a 6-layer CNN using 32 filters at each layer (see \Cref{fig::cf-conv-arch}) on the Cats vs. Dogs dataset with image sizes of $150 \times 150 \times 3$ \cite{asirra-a-captcha-that-exploits-interest-aligned-manual-image-categorization}.  Early experiments revealed difficulties in learning when the architecture operated solely in the Fourier domain. These challenges may stem from the complexities of differentiating through complex-valued activation functions \cite{8495012, 7458737}. To address this, we apply the (real) inverse FFT after each spectral convolutional, followed by traditional activation functions in the spatial domain, as illustrated in \Cref{fig::cf-conv-arch} (left). The outputs from the convolutional layers are average pooled across the channel dimension and fed into linear layers with 128 and 64 neurons, respectively, with ReLU activations applied throughout. A final sigmoid layer is used for binary classification.

To maintain parameter parity with the baseline $3 \times 3$ spatial CNN, we carefully select the architecture for CF-Convs. In the $\Phi_\Theta(H, W, C_{\text{in}}, C_{\text{out}})$ parameterization, which uses a single MLP, a network with [32, 32, 32, 16, 16, 16, 8, 8, 8, 1] neurons is employed. Although we advocate for this configuration, we also evaluate the $\Phi_\Theta(H, W)$ parameterization for comparison. This approach, with one MLP per $\{C_{\text{in}}, C_{\text{out}}\}$ kernel pair, uses only [2, 1] neurons per MLP. Both configurations use ReLU activations between linear layers. Our training protocol incorporates a straightforward augmentation pipeline involving up to $3$ augmentations per image using the \textit{RandAugment} class provided by \textit{KerasCV} \cite{wood2022kerascv}, which is exclusively applied to the training set.

\Cref{tab::sparse_params_accuracy} compares the performance of different models. The $\Phi_\Theta(H, W, C_{\text{in}}, C_{\text{out}})$ models with sparse updates outperform the $\Phi_\Theta(H, W)$ models, likely due to the increased complexity and expressiveness of the deeper MLP with more neurons. Additionally, the $\Phi_\Theta(H, W, C_{\text{in}}, C_{\text{out}})$ approach maintains a parameter count similar to the baseline $3 \times 3$ spatial CNN, while still achieving higher performance than the $\Phi_\Theta(H, W)$ configuration, which required more parameters. 

However, CF-Conv performance still trails $3 \times 3$ spatial CNNs.  This discrepancy suggests that CF-Convs may require more complex MLP architectures or further optimization to fully realize their potential. Moreover, the learning dynamics of spatial CNNs are well-understood and have been extensively validated empirically, whereas CF-Convs are still relatively new and may benefit from additional refinement and tuning.

We also examine the impact of different numbers of selected points on model performance, experimenting with $2^{12}$, $2^{15}$, and $2^{18}$ selected positions. The results indicate that smaller samples introduce more noise and provide less accurate gradient approximations. In contrast, using $2^{18}$ ($\sim$260,000) selected positions yields the best performance, striking an optimal balance between gradient approximation accuracy and memory usage. This configuration allows for stable and efficient training, while still fitting comfortably into GPU memory.

These results demonstrate that while CF-Convs have promising potential, further refinement is needed to match the performance of traditional spatial CNNs. Nonetheless, the flexibility and parameter efficiency of the $\Phi_\Theta(H, W, C_{\text{in}}, C_{\text{out}})$ configuration with sparse updates indicates that CF-Convs offer a viable path forward, especially for applications requiring larger and more complex architectures.
\section{Discussion}

Our proposed method for scaling CF-Conv networks via sparse kernel updates addresses key challenges in memory utilization and training speed. By introducing sparse updates, we significantly reduce the memory required during training, making the method feasible for large-scale applications. However, the technique still has limitations. Each CF-Conv layer must store a stateful kernel variable with dimensions $H \times W \times C_{\text{in}} \times C_{\text{out}} \times 2$, accounting for the real and imaginary components. For models with many convolutional filters or large spatial dimensions, this can still become memory-intensive. To mitigate this, model-parallelism could be a viable strategy, where model weights are distributed across multiple GPUs or TPUs to alleviate memory constraints without sacrificing training speed \cite{shoeybi2019megatron}.

Another challenge is that applying pointwise activation functions directly in the Fourier domain yields suboptimal performance in our experiments.  This may be a result of convolution in the Fourier domain (pointwise multiplication) and pointwise application of activation functions as both treat frequencies independently (pointwise operations).  This may prevent the network from capturing inter-frequency interactions, which could be crucial for complex tasks. To address this, we employ an inverse Fourier transform after each convolution, applying activations in the spatial domain. While this introduces additional computational complexity due to the repeated use of FFT and IFFT, it allows the network to maintain the benefits of learning in the Fourier domain while capturing essential inter-frequency interactions.

The development of improved complex-valued activation functions could provide an alternative solution. With improved complex-valued non-linearities, it may become possible to directly apply activations in the Fourier domain without the need for intermediate transforms. This would allow the network to fully utilize phase and amplitude information, making the approach particularly appealing for domains like audio, radar, and sonar image processing, where such information is critical.
\section{Conclusion}

Our work introduces CF-Convs, a novel approach for learning continuous convolutional kernels in the Fourier domain.  CF-Convs address three of the fundamental challenges of continuous convolutional kernel learning: parameter efficiency, memory efficiency and training speed. Additionally, our approach mitigates the spectral bias phenomena by learning directly in the spectral domain. This allows CF-Convs to capture a broader range of frequencies than its predecessors. While CF-Convs still lag behind traditional convolutions, our novel training algorithm allows for CF-Convs to learn convolutional kernels of arbitrary size, making them a promising direction for larger-scale applications.
{
    \small
    \bibliographystyle{ieeenat_fullname}
    \bibliography{main}
}


\end{document}